\def\year{2021}\relax
\documentclass[letterpaper]{article} % DO NOT CHANGE THIS
\usepackage{aaai21}  % DO NOT CHANGE THIS
\usepackage{times}  % DO NOT CHANGE THIS
\usepackage{helvet} % DO NOT CHANGE THIS
\usepackage{courier}  % DO NOT CHANGE THIS
\usepackage[hyphens]{url}  % DO NOT CHANGE THIS
\usepackage{graphicx} % DO NOT CHANGE THIS
\def\UrlFont{\rm}  % DO NOT CHANGE THIS
\usepackage{natbib}  % DO NOT CHANGE THIS AND DO NOT ADD ANY OPTIONS TO IT
\usepackage{caption} % DO NOT CHANGE THIS AND DO NOT ADD ANY OPTIONS TO IT
\title{Deep Repulsive Prototypes for Adversarial Robustness}
\author{
    Alex Serban\textsuperscript{\rm 1}, Erik Poll\textsuperscript{\rm 1}
    and Joost Visser\textsuperscript{\rm 2} \\
}

\affiliations{
    \textsuperscript{\rm 1}Radboud University\\
    \textsuperscript{\rm 2}Leiden University \\
    The Netherlands \\
    a.serban@cs.ru.nl
}

\title{Deep Repulsive Prototypes for Adversarial Robustness}
\usepackage{graphicx}
\usepackage{amsmath}
\usepackage{rotating}
\usepackage{cite}
\usepackage{threeparttable}
\usepackage{float}
\usepackage{enumitem}
\usepackage{graphicx}

\usepackage{array}
\newcolumntype{P}[1]{>{\centering\arraybackslash}p{#1}}
\newcolumntype{M}[1]{>{\centering\arraybackslash}m{#1}}
\newcolumntype{B}[1]{>{\centering\arraybackslash}b{#1}}
\usepackage{booktabs}
\graphicspath{{figures/}}
\DeclareGraphicsExtensions{.pdf,.jpeg,.png}
\usepackage{todonotes}
\usepackage{booktabs}
\usepackage{color,colortbl}
\definecolor{LGreen}{rgb}{.8,.9,.8}
\definecolor{LGray}{rgb}{.93,.93,.93}
\usepackage{pifont}
\usepackage{makecell}

\usepackage{adjustbox}

\usepackage{numprint}

\usepackage{acronym}
\usepackage{graphicx}
\usepackage{mathtools, amscd, mathrsfs}
\usepackage{multirow}
\usepackage{url}
\usepackage{tikz}
\usepackage{tikz-cd}

\usepackage{tabu}

\usetikzlibrary{shapes.geometric}
\usepackage{amsfonts}
\usepackage{multirow}
\usepackage{colortbl}
\usepackage{bm}
\usepackage{array}

\def\eqref#1{equation~\ref{#1}}
\def\1{\bm{1}}

\def\vtheta{{\bm{\theta}}}

\def\vx{{\bm{x}}}

\DeclareMathAlphabet{\mathsfit}{\encodingdefault}{\sfdefault}{m}{sl}
\SetMathAlphabet{\mathsfit}{bold}{\encodingdefault}{\sfdefault}{bx}{n}
\def\gF{{\mathcal{F}}}

\def\gS{{\mathcal{S}}}

\def\gU{{\mathcal{U}}}

\def\gX{{\mathcal{X}}}
\def\gY{{\mathcal{Y}}}
\def\gZ{{\mathcal{Z}}}

\def\sR{{\mathbb{R}}}

\DeclareMathOperator*{\argmin}{arg\,min}

\DeclareMathOperator{\sign}{sign}

\usepackage{todonotes}
\newboolean{showcomments}
\setboolean{showcomments}{true}
\ifthenelse{\boolean{showcomments}}
  {\newcommand{\nb}[2]{
  \fbox{\bfseries\sffamily\scriptsize#1}
    {\sf\small$>$\textit{#2}$<$}
   }
  }
  {\newcommand{\nb}[2]{}
   
  }
  
\definecolor{alexcolor}{rgb}{0.4,0.6,0.2}
\definecolor{carloscolor}{rgb}{0.2,0.6,0.6}
\definecolor{iliascolor}{rgb}{0.8,0.5,0.3}
\definecolor{zhourancolor}{rgb}{0.0,0.0,1}
\definecolor{todocolor}{rgb}{0.9,0.1,0.1}
\definecolor{changedcolor}{rgb}{0.42,0.27,0.57}

\newcommand{\crop}[1]{}

\newcommand{\specialcell}[2][c]{%
  \begin{tabular}[#1]{@{}c@{}}#2\end{tabular}}

\newcommand{\ie}{i.e.,}
\newcommand{\eg}{e.g.,}

\newcommand{\eqq}{Eq.}

\acrodef{ML}[ML]{machine learning}
\acrodef{DL}[DL]{deep learning}
\acrodef{DNN}[DNN]{deep neural network}
\acrodef{SVM}[SVM]{Support Vector Machine}
\acrodef{RL}[RL]{Reinforcement learning}
\acrodef{PAC}[PAC]{Probably Approximately Correct}

\acrodef{MILP}[MILP]{mixed integer linear programming}

\acrodef{BO}[BO]{bayesian optimization}
\acrodef{NES}[NES]{natural evolution strategies}

\acrodef{wrt}[\emph{w.r.t}]{with respect to}
\acrodef{st}[\emph{s.t.}]{such that}

\acrodef{fgsm}[FGSM]{Fast Gradient Sign Method}
\acrodef{bim}[BI]{Basic Iterative}
\acrodef{illcm}[ILC]{Iterative Least-likely Class}
\acrodef{jsma}[JSMA]{Jacobian-based Saliency Map Attack}
\acrodef{uap}[UAP]{Universal Adversarial Perturbations}
\acrodef{opa}[OPA]{One Pixel Attack}
\acrodef{pgd}[PGD]{projected gradient descent}
\acrodef{rssa}[RSSA]{Randomised Single Step Attack}
\acrodef{eat}[EAT]{Ensemble Adversarial Training}
\acrodef{gaa}[GAA]{Generative Adversarial Attacks}
\acrodef{gan}[GAN]{generative adversarial network}
\acrodef{nae}[NAE]{Natural Adversarial Examples}
\acrodef{atn}[ATN]{Adversarial Transformation Networks}
\acrodef{vae}[VAE]{Variational Auto-Encoders}
\acrodef{cfoa}[CFOA]{Complete First Order Adversary}
\acrodef{iid}[i.i.d]{independent and identically distributed}
\acrodef{bpda}[BPDA]{Backward Pass Differentiable Approximation}
\acrodef{alp}[ALP]{Adversarial Logit Pairing}
\acrodef{fbgan}[FB-GAN]{Featurized Bidirectional Generative Adversarial Networks}
\acrodef{sap}[SAP]{Stochastic Activation Pruning}
\acrodef{mat}[MAT]{Multi-strength Adversarial Training}
\acrodef{dam}[DAM]{Dense Associative Memory}
\acrodef{zoo}[ZOO]{Zeroth Order optimisation}
\acrodef{sa}[STA]{Strong Adversary}
\acrodef{lm}[LM]{Linear Models}
\acrodef{dt}[DT]{Decision Trees}
\acrodef{knn}[KNN]{K-nearest Neighbour}
\acrodef{ADAS}[ADAS]{Advanced Driver-Assistance Systems}
\acrodef{HW}[HW]{Hardwaere}
\acrodef{SW}[SW]{Software}
\acrodef{SDLC}[SDLC]{Software Development Life-Cycle}
\acrodef{AE}[AE]{Adversarial Example}

\allowdisplaybreaks

\usepackage{caption}
\usepackage{subcaption}

\usepackage{dirtree}

\usepackage{lscape}
\usepackage{afterpage}

\definecolor{alexcolor}{rgb}{0.4,0.6,0.2}
\definecolor{erikcolor}{rgb}{0.2,0.6,0.6}
\definecolor{joostcolor}{rgb}{0.8,0.5,0.3}

\usepackage[switch]{lineno}  %

\begin{document}
\maketitle
% \linenumbers  %

\begin{abstract}
    While many defences against adversarial examples have been proposed, finding robust machine learning models is still an open problem.
    The most compelling defence to date is adversarial training and consists of complementing the training data set with adversarial examples.
    Yet adversarial training severely impacts training time and depends on finding representative adversarial samples.
    In this paper we propose to train models on output spaces with large class separation
    in order to gain robustness without adversarial training.
    % By doing so, models gain robustness to adversarial examples without adversarial training.
    We introduce a method to partition the output space into class prototypes with large separation and train models to preserve it.
    % enforce large class separation on the output space and train models to preserve it.
    Experimental results shows that models trained with these prototypes~--~which we call deep repulsive prototypes~--~gain robustness competitive with adversarial training, while also preserving more accuracy on natural samples. 
    Moreover, the models are more resilient to large perturbation sizes.
    For example, we obtained over 50\% robustness for CIFAR-10, with 92\% accuracy on natural samples and over 20\% robustness for CIFAR-100, with 71\% accuracy on natural samples without adversarial training. 
    For both data sets, the models preserved robustness against large perturbations better than adversarially trained models.
    % On both data sets, the models preserve more than 40\% of robustness for very large perturbations.
\end{abstract}

\section{Introduction}
\label{sec:intro}

Although a plethora of adversarial defences have been proposed~--~ranging from input projections~\cite{guo2017countering} to formal guarantees that no adversarial examples can be found within some bounds~\cite{huang2018safety}~--~robustness to adversarial examples remains an open problem.

The most compelling defence to date is adversarial training and consists of complementing the training data set with adversarial samples~\cite{goodfellow2014explaining}, or training only on perturbed data~\cite{madry2017towards}.
Yet adversarial training is subject to several trade-offs. % detriments?
Firstly, the time needed to generate adversarial examples substantially increases training time.
Recent attempts to generate adversarial examples faster exist~\cite{wong2020fast}.
However, they are (at the moment) unstable and introduce new issues such as catastrophic forgetting~\cite{andriushchenko2020understanding}. 

Secondly, a trade-off between accuracy on natural samples and robustness on adversarial examples is known to exist~\cite{zhang2019theoretically}.
This trade-off implies that robustness against adversarial examples comes with a cost of losing accuracy on natural examples, and can be controlled through adversarial training~\cite{zhang2019theoretically}.
Lastly, adversarial training overfits on training data and provides little robustness against data outside this distribution~~\cite{zhang2019limitations,rice2020overfitting}.

A model robust to adversarial examples should provide: (i)~inter-class separability, (ii)~intra-class compactness, and (iii)~marginalisation or removal of non-robust features~\cite{smith2019useful,ilyas2019adversarial}.
However, adversarial training does not impose explicit constraints for meeting these properties (\eg~by specifying inductive biases).
Therefore, it depends only on finding representative adversarial examples for training.

In supervised classification, adversarial training uses the standard softmax cross-entropy loss.
Recently, there is increasing evidence that softmax partitions the output space into class centroids situated at equal distance from the origin (inter-class separability), and that adversarial robustness can be improved by clustering the data points in the proximity of these centroids (intra-class compactness)~\cite{hess2020softmax,papernot2018deep}.

However, the distance between class centroids is insufficient to provide robustness, and even models with high intra-class compactness are vulnerable to adversarial attacks.
In this paper we tackle this issue by enforcing large inter-class separation prior to training using class prototypes~\cite{snell2017prototypical}.
By making use of this inductive bias we gain more control over the output space structure, and can decrease the number of training samples needed~\cite{mettes2019hyperspherical}.
% This inductive bias allows more control over the output space, and takes a more practical path to  robustness than learning without inductive biases. %, as in adversarial training, 
% By making use of this inductive bias we gain more control over the output space, and provide a practical alternative for robustness than learning without inductive biases.
%~\cite{mettes2019hyperspherical}.
% and decreases the number of training samples needed~\cite{mettes2019hyperspherical}.
% TODO: Add a point that explicit inductive biases help
% robustness
% make everything more practical; more transferable from lab to practice
% more interpretable 

We show that training with class prototypes optimised to provide large inter-class separation helps to gain robustness competitive with adversarial training, \emph{without} adversarial training.
Moreover, training with class prototypes involves a smaller trade-off between accuracy and robustness, and a higher resilience against large perturbations.
The prototypes are built prior to training with little overhead, through an optimisation procedure that increases the distance between their centres. % 
As a result of this repelling optimisation procedure, and because we use deep neural networks for empirical validation, we call the prototypes \emph{deep repulsive prototypes}.
We test repulsive prototypes on CIFAR-10 and CIFAR-100, and observe consistent results on both data sets, with 51.3\% and 20.5\% robustness against iterative adversarial attacks.

The rest of the paper is organised as follows. 
Initially, we introduce background information and discuss related work (Section~\ref{sec:background}).
Later, we present repulsive prototypes (Section~\ref{sec:proto}), followed by an evaluation against white and black-box attacks (Section~\ref{sec:exp}).
We conclude with a discussion (Section~\ref{sec:discussion}) and future work (Section~\ref{sec:conclusion}).

\section{Background and Related Work}
\label{sec:background}
We focus on supervised classification, \ie~given a set of inputs sampled from $\gX$ and corresponding labels sampled from $\gY$, an
algorithm finds a mapping from $f(\cdot, \vtheta): \gX \rightarrow \gY$ which minimises the number of misclassified inputs.
We assume that $\gX$ is a metric space such that a distance function $d(\cdot)$ between two points of the space exists.
The error made by a prediction $f(\vx) = \hat{y}$ when the true label is $y$ is measured by a loss function $l:\gY \times \gY \rightarrow \sR$ with non-negative values when the labels are different and zero otherwise.
$f(\cdot)$ is defined over a hypotheses space $\gF$ which encompasses any mapping from $\gX$ to $\gY$ and can take any form~--~\eg~a linear function or a neural network.
Through learning, an 
% \ac{ML} 
algorithm selects $f^{*}(\cdot)$ from $\gF$ such that the expected empirical loss on a training data set $\gS$ consisting of pairs of samples $(\vx_i, y_i) \sim \gZ = \gX \times \gY$ is minimal.

A robust solution to the minimisation problem above involves immunising it against uncertainties in the input space.
In the adversarial examples setting, uncertainties are modelled in the space around an input $\vx'$: $\gU_{\vx} = \{\vx' | d(\vx, \vx') \leq \epsilon\}$. %, $d(\cdot)$ is a distance function defined on the metric space $\gX$.
The robust counterpart of the learning problem becomes:
\begin{equation}
f = \argmin_{f \in \gF} \mathbb{E}_{(\vx, y)  \sim \gS}  [ \max_{\vx' \in \gU_{\vx}} l(\vtheta, \vx', y)],
\label{eq:rerm}
\end{equation}
\noindent
where $\vx'$ is a realization of $\vx$ in the uncertainty set described by $d(\cdot)$.
Common distance functions $d(\cdot)$ are the Euclidean $L_2$ distance or the Chebyshev $L_\infty$ distance.
We further use the term robustness to define the accuracy of a model on adversarial examples, $\mathbb{E}_{(\vx, y)  \sim \gS}  [f(\vx') = y]$.

Adversarial training consists of adding a regularization term to the loss function:
\begin{equation}
\tilde{l}(\cdot) =    \alpha  l(\vtheta, \vx, y) +  (1-\alpha)  l(\vtheta, \vx', y),
\label{eq:adv_training}
\end{equation}
\noindent
where $\vx'$ is an adversarial example generated from input $\vx$ and $\alpha$ controls the contribution of adversarial examples to the loss.
The most effective choice of $\alpha$ is zero~\cite{madry2017towards}, which poses~\eqq~(\ref{eq:rerm}) as a min-max problem where the inner maximisation problem seeks to find the worst adversarial example for an input and the outer minimisation problem seeks to strengthen the model against it.

Finding an exact solution to the inner maximisation problem is not feasible for adversarial training.
The most effective approximation solution relies on iteratively taking small steps towards maximising the loss function, and projecting the outcome on the space defined by  $d(\cdot)$; a procedure called \ac{pgd} and defined as follows:
\begin{equation}
\vx_{n}' = \prod_{\vx+\gU_\vx} (\vx_{n-1}' + \epsilon \sign( \nabla_{\vx'_{n-1}} l(\vtheta, \vx'_{n-1}, y)) ) .
\label{eq:pgd}
\end{equation}
The quality of the adversarial examples depends on the number of iterations $n$. 
Using a large number of iterations, \ac{pgd} can better approximate the space around an input we want to provide robustness to, and leads to more representative adversarial examples for training.
However, it negatively affects training time.
When $n=1$, the method is called the \ac{fgsm}.

Several attempts have been made to change the training procedure in order to enforce inter-class separability or intra-class compactness, and also decrease the impact of adversarial training.
\citet{mao2019metric}~used a triplet loss (inspired by metric learning), where one element of the triplet loss is an adversarial example.
An attempt to reduce the impact of adversarial training on training time was made by only generating one adversarial example for each triplet data.
Further inductive biases, such as careful negative sample selection for the triplet loss, help to improve robustness.

\citet{papernot2018deep} showed that explicitly tailoring intra-class compactness using k-neighbours in the representation space helps to detect adversarial examples.
\citet{hess2020softmax} proved a similar result and proposed a method based on the Gauss kernel to enforce intra-class compactness and improve robustness.
\citet{pang2019rethinking} introduced a loss function to enforces inter-class separability using the centroids of the Max-Mahalanobis distribution. 
During inference, the class centroid closer to the input's deep representation (measured using the Euclidean distance) was used to classify an input.
The defence builds on earlier work by~\citet{pang2019mixup}, where at inference time an input is interpolated with samples from the same predicted class, and from distinct classes in order to alleviate the impact of perturbations.
Unfortunately, none of these defences proved effective~\cite{tramer2020adaptive}.

\citet{jin2020manifold} showed that manifold regularisation improves adversarial robustness significantly while retaining better accuracy on natural examples, without adversarial training.
Their proposal induces local stability in the neighbourhood of natural inputs even if the model classifies the inputs incorrectly.
This is in contrast with adversarial training, where a model is trained to classify correctly worst case adversarial examples.

\citet{mustafa2019adversarial}~used class prototypes to enforce inter-class separability, by including a prototype separation constraint in the loss function.
A convex polytope is assigned as prototype to each class and during training the distance between all class polytopes is maximised.
Thus the class centroids are learned together with the internal representation.
However, adding the distance maximisation term to the loss function does not suffice to improve robustness, and they propose to add similar constraints to hidden layers.
When paired with adversarial training, robustness increases at a decreased cost for accuracy on natural samples. 

\citet{mettes2019hyperspherical} showed that defining class prototypes a priori to training in order to enforce desired properties of the output space (\eg~large margin separation) improves training in several settings; such as few-shot classification, regression or joint classification and regression.
Instead of constantly re-estimating and calibrating the prototypes~--~as in~\citet{mustafa2019adversarial} or others~\cite{snell2017prototypical,guerriero2018deep}~--~they propose to define the output space as a hyper-sphere and partition it into predefined class prototypes.
During training, the distance between the model's output and class prototypes is minimised.
The a priori definition of class prototypes enables control over several factors such as the output space size, or its shape.
In this paper we take a similar path and define class prototype prior to training. %, instead of learning them as in~\cite{mustafa2019adversarial}.

\section{Repulsive Prototypes for Robustness}
\label{sec:proto}

\begin{figure}[t]
    \centering
    \includegraphics[width=6.5cm, keepaspectratio]{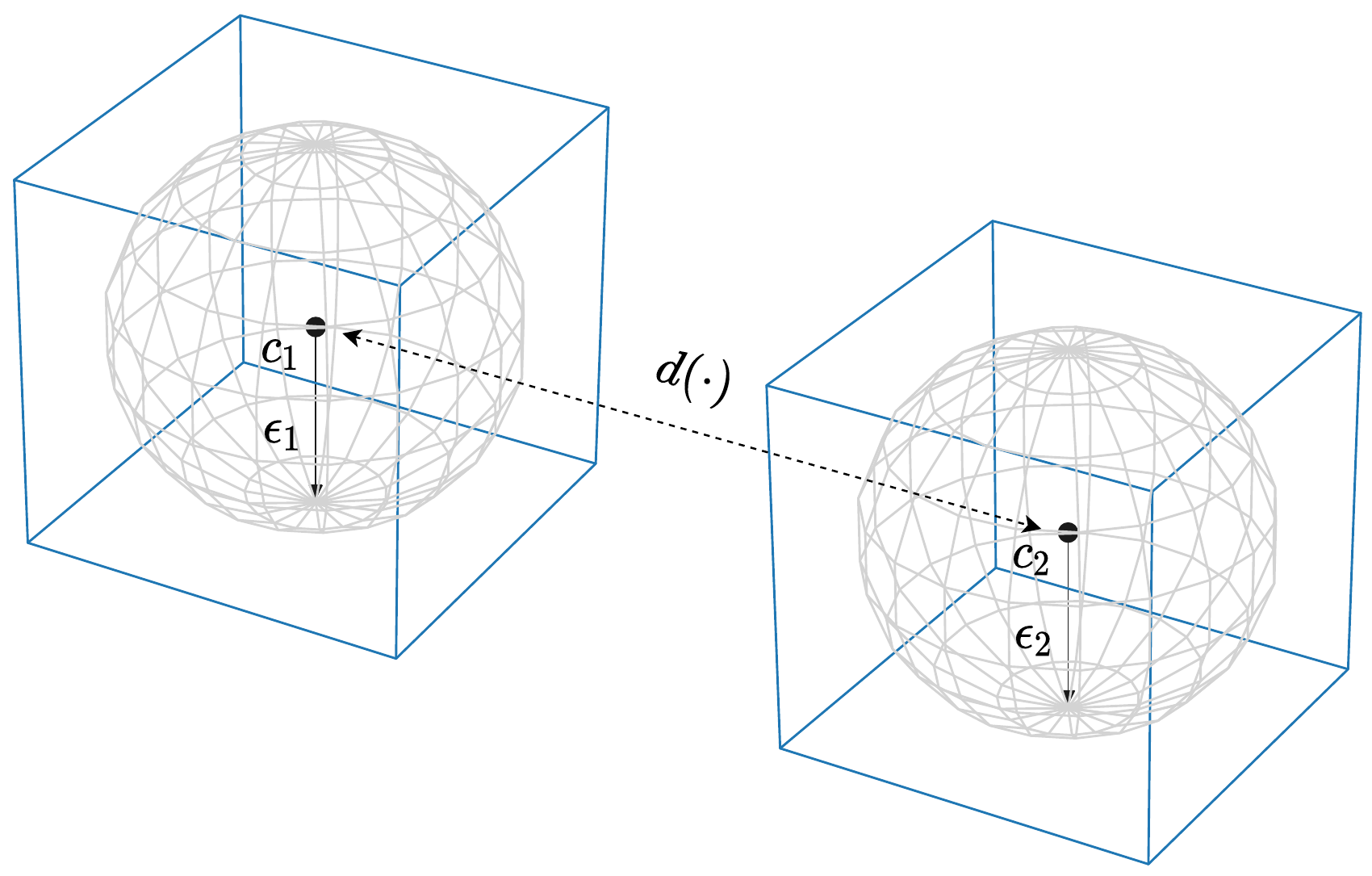}
    \caption{Repulsive prototypes. The grey spheres represent prototypes obtained using the $L_2$ distance, and the blue cubes are obtained using the $L_\infty$ distance.}
    \label{fig:repulsive_proto}
\end{figure}

The idea behind building class prototypes for adversarial robustness is to explicitly design prototypes with large inter-class separation, and during training enforce intra-class compactness.
To this end, the input or the output space is partitioned into hyper-planes specific to each class, to which we impose separation constraints.
Similar approaches have been used in the past, \eg~by~\citet{schiilkop1995extracting} who used the smallest sphere enclosing the data to estimate the VC-dimension for support vector classifiers, or by~\citet{wang2005pattern} who used separating spheres in the feature space for classification.

We propose an approach similar to~\citet{wang2005pattern} and \citet{nguyen2015repulsive}, and more recent work by~\citet{mettes2019hyperspherical}, where the separation boundaries are imposed to the output space~--~instead of the input space~--~because the output space allows more flexibility and can achieve larger margin separation.
Prior to learning, the $D$-dimensional output space is divided into $k$ prototypes $P=\{p_1, \cdots, p_k\}$, where each prototype corresponds to a class.
For a binary classification problem and a Euclidean output space, we wish to find two hyper-spheres with centres at $c_1, c_2$~--~one enclosing samples from the positive class and the other enclosing samples from the negative class~--~and maximise the distance between them: 

\begin{equation}
	\begin{aligned}
	    &\min_{\epsilon_1, \epsilon_2, c_1, c_2} & & \epsilon_1^2 + \epsilon_2^2 - r \| c_1 - c_2 \|^2 \\ 
    	&  s.t.  &  & \| f(\vx_i) - c_1 \|^2 \leq \epsilon_1^2, \forall i, y_i = +1  \\
    	&    &  & \| f(\vx_i) - c_1 \|^2 \geq \epsilon_1^2, \forall i, y_i = -1  \\
    	&    &  & \| f(\vx_i) - c_2 \|^2 \leq \epsilon_2^2, \forall i, y_i = -1  \\
    	&    &  & \| f(\vx_i) - c_2 \|^2 \geq \epsilon_2^2, \forall i, y_i = +1, 
%		& & & f(\vx) = y, \\
%		& & & \hat{y} \neq y,		
	\end{aligned}
    \label{eq:proto_constrained}
\end{equation}
where $r$ is a constant that represents the repulsive degree between the two prototypes, and $\epsilon_1, \epsilon_2$ define the $d(\cdot)$-ball around the prototype centres for which  we want to provide robustness (corresponding to the uncertainty set in \eqq~(\ref{eq:rerm})).

For non-separable data sets, the constraints above can be relaxed by introducing slack variables and regularisation terms to the objective function.
Although the objective in \eqq~(\ref{eq:proto_constrained}) is not convex, it can be reformulated to have a convex form and solved using Lagrange multipliers.
However, in practice the constraints can be relaxed and the problem can be solved in two steps: firstly find prototypes with large separation (to provide inter-class separability), and secondly train models to fit the data within the proximity of the prototype centres (to provide intra-class compactness).
An approximate solution to the first problem can be found by running gradient descent on the unconstrained objective:
\begin{equation*}
	\begin{aligned}
	\label{eq:proto_generic}
	    &\min_{\epsilon_1, \epsilon_2, c_1, c_2} & & \epsilon_1^2 + \epsilon_2^2 - r \| c_1 - c_2 \|^2,
	\end{aligned}
\end{equation*}
with a generalisation to k-classes and any metric space:
\begin{equation}
    \begin{aligned}
    \label{eq:proto}
        &\min_{\epsilon, c} & & \epsilon^k - r\sum_{(i,j, i \neq j) \in k} d(c_i, c_j).
    \end{aligned}
\end{equation}
The choice of $r$ can also be controlled using the learning rate $\mu$ for gradient descent.
The choice of $d(\cdot)$ influences the prototypes and the classification regions defined in the output space.
For example, using the Euclidean $L_2$ distance leads to hyper-spherical classification regions, and using the Chebyshev $L_\infty$ distance leads to hyper-cubical regions.
An illustration is provided in Figure~\ref{fig:repulsive_proto}, where the grey spheres represent regions for perturbations in the $L_2$ space and the blue cubes are regions for perturbations in the $L_\infty$ space around the centres. % $c_1, c_2$.
Iterating over \eqq~(\ref{eq:proto}) is equivalent to increasing $d(\cdot$) or adding slack variables to \eqq~(\ref{eq:proto_constrained}).
Larger distances between class prototypes introduce buffers between classification boundaries and should improve robustness.

The second step~--~training models to fit the data within the proximity of the prototype centres~--~can be solved by minimising the distance between the prototype centres and the model's output.
The choice for this distance function is part of the threat model and it is the same as $d(\cdot)$ from  \eqq~(\ref{eq:proto}), which induces the following loss function:
\begin{equation}
    % l = \mathbb{E}_{(\vx, y)  \sim \gS}~ [1 - d(f(\vx), p_y])^2,
    l = \sum_{i=1}^{N} (1 - d(f(\vx_i), p_{y_i}))^2,
    \label{eq:loss}
\end{equation}
where $p_{y_i}$ is the prototype specific to class $y_i$.

\section{Empirical Evaluation}
\label{sec:exp}

All experiments are performed using a vanilla ResNet-18 network (the smallest variant of ResNet). 
Capacity is known to help adversarial robustness~\cite{madry2017towards,xie2020smooth}. 
Therefore, we avoid using larger networks.
During training with repulsive prototypes only natural samples are used, \ie~\emph{no} adversarial training is performed.

Firstly, we adopt a white-box threat model for testing, where attackers presumably have full knowledge of the model under attack, the training and the testing data~\cite{carlini2019evaluating}.
To generate adversarial examples we use the \ac{pgd} (\eqq~(\ref{eq:pgd})) PyTorch implementation from Cleverhans, with different iterations and random restarts~\cite{papernot2016technical}.
Testing against larger $n$ is recommended, as it shows if the model exhibits a false sense of robustness or obfuscates attack vectors~\cite{carlini2019evaluating}.

Attackers are constrained to generate adversarial examples in the $\epsilon=8$ (normalised) $L_p$ norm ball around inputs~--~a common benchmark for adversarial robustness.
Since the distance between prototypes is larger than $\epsilon$, we expect models trained with repulsive prototypes to also exhibit resilience to higher $\epsilon$ values.
In order to test this hypothesis, we use robustness curves obtained by step-wise increasing the size of the perturbation in the interval $[8, 16]$.

We compare with results from literature on two common data sets; CIFAR-10 and CIFAR-100~\cite{krizhevsky2009learning}.
The first one consists of \numprint{60000} 32x32 colour images and 10 classes (with \numprint{5000} images for training and \numprint{1000} images for testing per class).
The second data set consists of \numprint{60000} 32x32 images and 100 classes (with \numprint{500} training images for training and \numprint{100} images for testing per class).
We use minimal data pre-processing for training, consisting of random cropping and random horizontal flip. % and normalisation.
No data pre-processing is used for testing.

For training, we use the cyclical learning rate~\cite{smith2017cyclical}, mixed precision arithmetic and early stopping, as they are reported to improve training time and prevent  overfitting~\cite{wong2020fast, rice2020overfitting}.
% All hyper-parameters and implementation details are available in the project's public repository\footnote{link suppressed for review}.

Later in this section we also adopt a black-box threat model, where attackers can only observe the outcome of the models under attack.
For evaluation we use the transferability attack, where adversarial examples are generated with a distinct model, and transferred to the models under attack~\cite{su2018robustness, carlini2019evaluating}.

\begin{table}[t]
    \centering
    \begin{tabular}{c|c|c|c}
    \specialcell{Output \\ Dimension ($D$)} & Epochs &  \specialcell{Clean \\ Samples} & PGD-20  \\
        \toprule
        % 10  & 50 & 81.5   & 26.7    \\
        % 20  & 50 & 90.3   & 33.5    \\
        50  & 50 & 90.3   & 37.2    \\
        100 & 50 & 90.6   & 39.9    \\
        200 & 50 & 89.5   & 40.7    \\
        \midrule
        100  & 100 & 91.0  & 48.7   \\
        200  & 100 & 91.1  &  38.7  \\
        \bottomrule
    \end{tabular}
    \caption{Prototype selection on CIFAR-10.}    
    \label{tab:protoselect}
\end{table}
\subsection{Prototype Selection}

Several parameters influence the quality of the prototypes: the output space dimension $D$, the choice of $\epsilon$, $r$, the learning rate $\mu$ and the number of epochs for solving \eqq~(\ref{eq:proto}).
Since $r$ and $\mu$ can be compressed to one constant, and (to some extent) the effect of the learning rate can be attenuated by running the optimisation longer, the most important parameters are the output space dimension and the number of epochs.
As mentioned earlier, when not mentioned otherwise we use the same choice for $\epsilon=8$ (normalised).

Previously, it has been shown that increasing $D$ can benefit both classification and regression~\cite{mettes2019hyperspherical}.
In order to determine the influence of $D$ on robustness and accuracy, we run an experiment on the CIFAR-10 data set, training a ResNet-18 model for 50 epochs with different output sizes $D \in \{ 50 ,100 ,200\}$~--~corresponding to multiplying the number of classes $k$ with factors of $\{5, 10, 20\}$.
Testing is performed with natural and perturbed samples using the \ac{pgd} attack, with $n=20$.
In all cases, the prototypes are generated by running gradient descent on \eqq~(\ref{eq:proto}) for 100 epochs, with $\mu=0.01$.
This modest optimisation budget is sufficient to obtain large distances between the prototypes.

The results are presented in Table~\ref{tab:protoselect}.
We observe that increasing the output dimension $D$ has almost no impact on accuracy on natural samples, but a significant impact on robustness, for all values of $D$ except the last one.
For the last two values of $D$ we ran training longer and observe that the largest output space (\ie~200) has a bigger tendency to overfit for adversarial examples, while maintaining similar accuracy on natural samples (corresponding to 100 epochs in Table~\ref{tab:protoselect}).
This phenomenon will be elaborated in Section~\ref{sec:discussion}.

The initial experiments on prototype selection reveal that the output size is important for adversarial robustness, but plays a marginal role for accuracy on natural samples.

\subsection{CIFAR-10}

\begin{figure}[t]
    \centering
    \includegraphics[width=7.5cm, keepaspectratio]{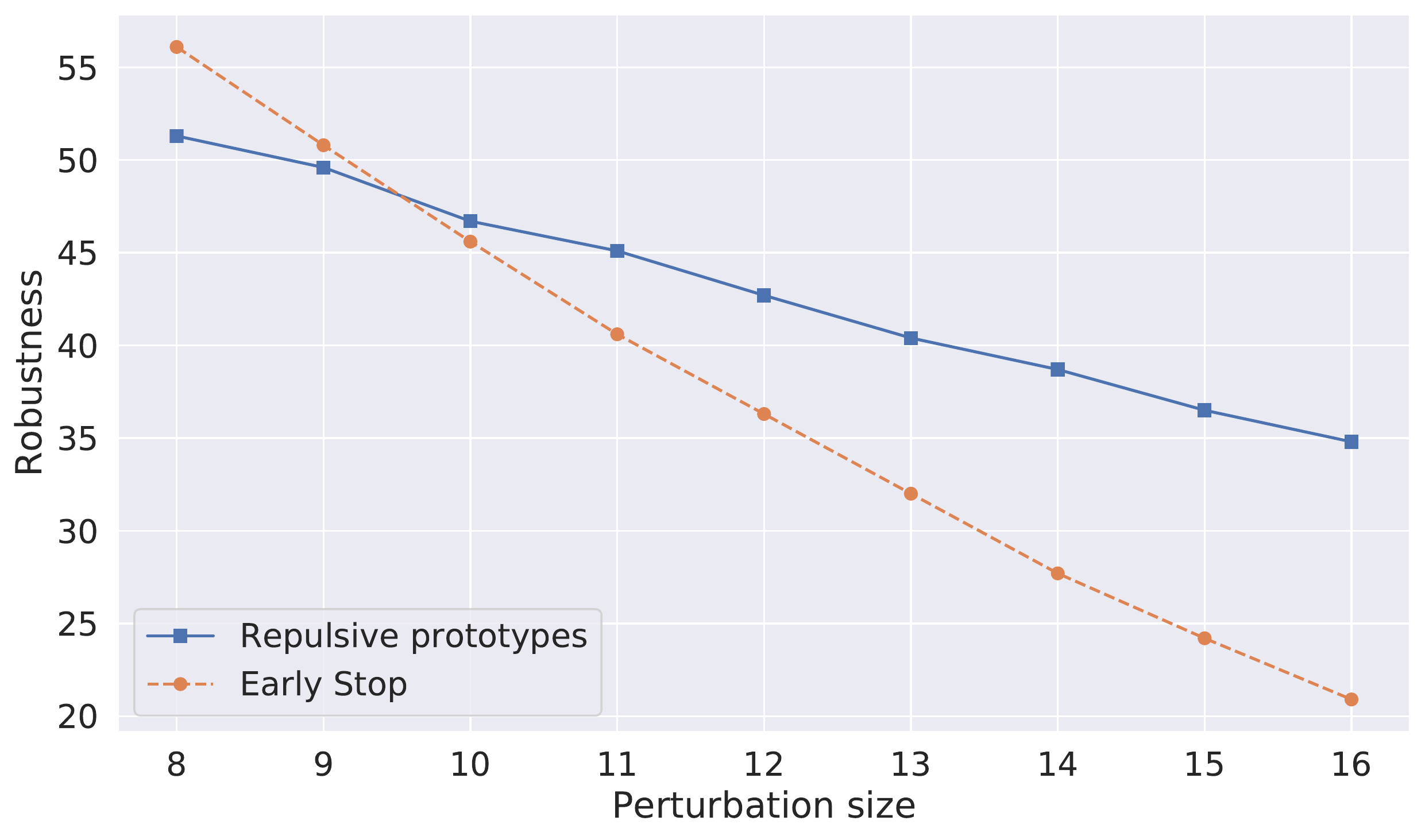}
    \caption{Robustness curves for CIFAR-10, obtained by testing with PGD-20, and various perturbation sizes ($\epsilon$).}
    \label{fig:cifar10_curve}
\end{figure}

Following the previous experiments, we present the results from training a ResNet-18 model on CIFAR-10 with the same parameters as earlier, but run the optimisation for longer and test it against stronger attacks.
For all experiments, the output dimension is $D=100$, corresponding to multiplying the number of classes by a factor of ten.

We benchmark our results against the following results from literature: (i) the initial results for adversarial training from~\citet{madry2017towards}, (ii) the improved results for adversarial training from~\citet{rice2020overfitting} which use early stopping to prevent overfitting in adversarial training, (iii) the work of~\citet{zhang2019theoretically} which trades more accuracy on natural samples in order to gain robustness, and (iv) the work of~\citet{mustafa2019adversarial} which use class prototypes jointly optimised during training, and where the inter-class separation constraints are applied to multiple layers, and paired with adversarial training.
We note that~\citeauthor{zhang2019theoretically} report the highest robustness.
However, \citeauthor{rice2020overfitting} showed that early stopping improves robustness, and reduces the gap between~\citeauthor{madry2017towards} and \citeauthor{zhang2019theoretically}, while also preserving more accuracy on natural samples.

While \citeauthor{madry2017towards} and \citeauthor{mustafa2019adversarial} use \ac{pgd} adversarial training with $n=7$, \citeauthor{zhang2019theoretically} and \citeauthor{rice2020overfitting} use $n=10$.
As mentioned above, adversarial training adds a non-trivial overhead, and a higher $n$ further increases it.
While faster methods to perform adversarial training exist, they achieve \emph{at most} similar results to classical adversarial training.
Therefore, we compare our results with the state-of-the-art for classical adversarial training.
Since our method does not add any significant overhead to training, whenever we discuss the impact on training we compare with~\citet{wong2020fast}, which is (at the moment) the fastest way to perform adversarial training, albeit not stable~\cite{andriushchenko2020understanding}.

The results are presented in Table~\ref{tab:cifar10}, where the acronyms follow the order above: (i) Madry~\cite{madry2017towards}, (ii) Early Stop~\cite{rice2020overfitting}, (iii) TRADES~\cite{zhang2019theoretically}, (iv) RHS~\cite{mustafa2019adversarial}. 
The Regular run was trained on natural samples with the softmax cross-entropy loss, and a multi-step learning rate scheduler that starts from $0.1$ and decays by a factor of $0.1$ at epochs 50 and 100.
For the models in literature we present the reported results, since with the exception of~\citeauthor{rice2020overfitting} the results could not be reproduced precisely.

\begin{table}[t]
    \centering
    \begin{tabular}{l|c|c|c|c|l}
        Run & Ep. &  Natural & \specialcell{PGD \\ 20} &  \specialcell{PGD \\ 100}  & \specialcell{Adv. \\ Training}  \\
        \toprule
        Regular\textsuperscript{1} & 120 & 93.7 &   0  &   0   & None \\

        % ResNet 18 Proto-100 & 90 &  91.2 \%   & 47.1 \% & 41.8 \% & 41.2 \% & - \\  
        
        % ResNet 18 Proto-100 & 110 & 90.7 \%  & 49.9 \% &  42.5 \% &  40.1 \% & - \\
        
        Repulsive\textsuperscript{1} & 127 & 92.0  &  51.3  &  48.4 & None \\      

        % ResNet 18 Proto-100 & 147 & 92.1 \%  &  49.7 \% &  40.0 \% &  36.4 \% & - \\      
        
        \midrule
        
        Madry\textsuperscript{2} & 200 & 87.2  & 45.8  & - &  PGD-7 \\
        
        Early Stop\textsuperscript{3} & 100 &  86.1 & 56.1  & - &  PGD-10 \\
        
        TRADES\textsuperscript{3} & 100 & 84.9 & 56.6 & - &   PGD-10 \\ 
        
        RHS\textsuperscript{4} & 300 & 91.8  & 42.6  &  - &    PGD-7 \\ 
        % Restricting Hidden Space        
        
        \bottomrule
    \end{tabular}
    \caption{CIFAR-10 results, $\epsilon=8$. The models are~\textsuperscript{1}ResNet-18,~\textsuperscript{2}PreActResNet-18,~\textsuperscript{3}WideResNet-34-10,~\textsuperscript{4}ResNet-110. Ep. abbreviates the number of training epochs.}    
    \label{tab:cifar10}
\end{table}

For the model trained using repulsive prototypes (the Repulsive run) we report the robustness against the \ac{pgd} attack with 20 and 100 iterations.
During training, the cyclical learning rate was reduced by a factor of ten compared to~\citet{smith2017cyclical}.
We found that using smaller learning rates benefits robustness and has little impact on natural accuracy.
The reason for this is that larger updates may push the samples closer to the decision boundaries, where it is easier for adversarial perturbations to induce undesirable behaviour.
For all models we also report the accuracy on natural samples, the number of epochs needed to reach the results and the architecture used for training. 

We observe that training with repulsive prototypes yields higher accuracy on natural samples (92\%) than methods based on adversarial training, and competitive robustness (51.3\%) compared with the state-of-the-art (56.6\%), at a relatively small increase of training epochs (+27).
This is a gain even for~\citet{wong2020fast}, which uses the \ac{fgsm} attack and thus requires at least two forward and backward passes at each epoch.
Moreover, TRADES and Early Stop use $n=10$ for adversarial training (which increases robustness over Madry), and use a WideResNet-34-10 architecture, which has over $34*10^6$ more training parameters than ResNet-18.
Both capacity and a higher $n$ are known to increase robustness~\cite{madry2017towards}.

Figure~\ref{fig:cifar10_curve} illustrates the robustness curve obtained by testing the models with \ac{pgd} $n=20$, against different perturbation sizes.
For comparison, we use the Early Stop model by~\citeauthor{rice2020overfitting}, the only one for which the results could be reproduced with precise accuracy.
We observe that training with repulsive prototypes yields models which are more resilient to higher perturbations than adversarially trained models (equivalent to a milder slope in Figure~\ref{fig:cifar10_curve}).
Moreover, the overall decrease in accuracy is significantly smaller for models trained with repulsive prototypes; preserving more than 60\% of the initial robustness when the perturbation size is doubled.

\subsection{CIFAR-100}
\begin{table}[t]
    \centering
    \begin{tabular}{l|c|c|c|c|l}
        Run & Ep. &  Natural & \specialcell{PGD \\ 20} &  \specialcell{PGD \\ 100}  & \specialcell{Adv. \\ Training}  \\
        \toprule
        Regular\textsuperscript{1} & 120 & 73.8 &   0  &   0   & None \\

        % Proto\textsuperscript{1} & 97 & 71.2 &  20.2  &   & None \\      

        Repulsive\textsuperscript{1} & 106 & 71.7 &  20.5  &  20.0  & None \\

        % ResNet 18 Proto-100 & 147 & 92.1 \%  &  49.7 \% &  40.0 \% &  36.4 \% & - \\      
        
        \midrule
        
        Madry\textsuperscript{2} & 200 &  59.8 & 22.6 & - & PGD-7 \\
        
        Early Stop\textsuperscript{2} & 100  & 52.7 &  28.1 & - & PGD-10 \\
        
        Early Stop-R\textsuperscript{2} & 100  & 54.1 &  20.8 & - & PGD-10 \\
        
        % TRADES\textsuperscript{4} & &  &  & - & PGD-10 \\ 
        
        RHS\textsuperscript{3} & 300 & 68.3  & 20.2  &  - &    PGD-7 \\ 
        
        \bottomrule
    \end{tabular}
    \caption{CIFAR-100 results, $\epsilon=8$. The models are are \textsuperscript{1}ResNet-18, \textsuperscript{2}PreActResNet-18, \textsuperscript{3}ResNet-110. Ep. abbreviates the number of training epochs.}    
    \label{tab:cifar100}
\end{table}

We perform and report complementary experiments on the CIFAR-100 data set.
The key difference between the two is that the number of classes increases by a factor of ten.
Therefore, the output space partitioning is more challenging.

Moreover, since the last fully connected layer of ResNet-18 has \numprint{512} nodes, we use a multiplicative factor of 50 instead of 100 for $D$ in order to preserve a possible compression in the last layer, as for CIFAR-10.
Experiments with different multiplicative factors, as those discussed in Table~\ref{tab:protoselect}, are available in the project's repository.

The results are presented in Table~\ref{tab:cifar100}, with the notable difference that TRADES was not tested on this data set neither in the original paper~\cite{zhang2019theoretically} or in the Early Stop paper~\cite{rice2020overfitting}.
Moreover, for Early Stop we could not reproduce the results reported in the paper.
Therefore, we also report on a new benchmark, Early Stop-R, which is obtained using the model parameters shared in the project's repository by~\citeauthor{rice2020overfitting}.
Also note that for CIFAR-100 Early Stop uses the PreActResNet-18 architecture instead of WideResNet.

We observe that training with repulsive prototypes yields significantly higher accuracy on natural samples (71.7\%) compared with adversarial training methods, where the maximum is achieved by~\citet{madry2017towards} (59.8\%).
Moreover, competitive robustness (20.5\%) with adversarial training (22.6\%) can be observed, at almost no increase in training epochs (+6).
The results reported for Early Stop by~\citeauthor{rice2020overfitting} show 7.6\% more robustness than training with repulsive prototypes, at the cost of losing 17\% accuracy on natural samples.
A similar result can be observed for CIFAR-10, which indicates that training with repulsive prototypes trades less accuracy on natural samples, at the cost of a modest contraction in robustness.
% This topic will be further elaborated in Section~\ref{sec:discussion}.

\begin{figure}[t]
    \centering
    \includegraphics[width=7.5cm, keepaspectratio]{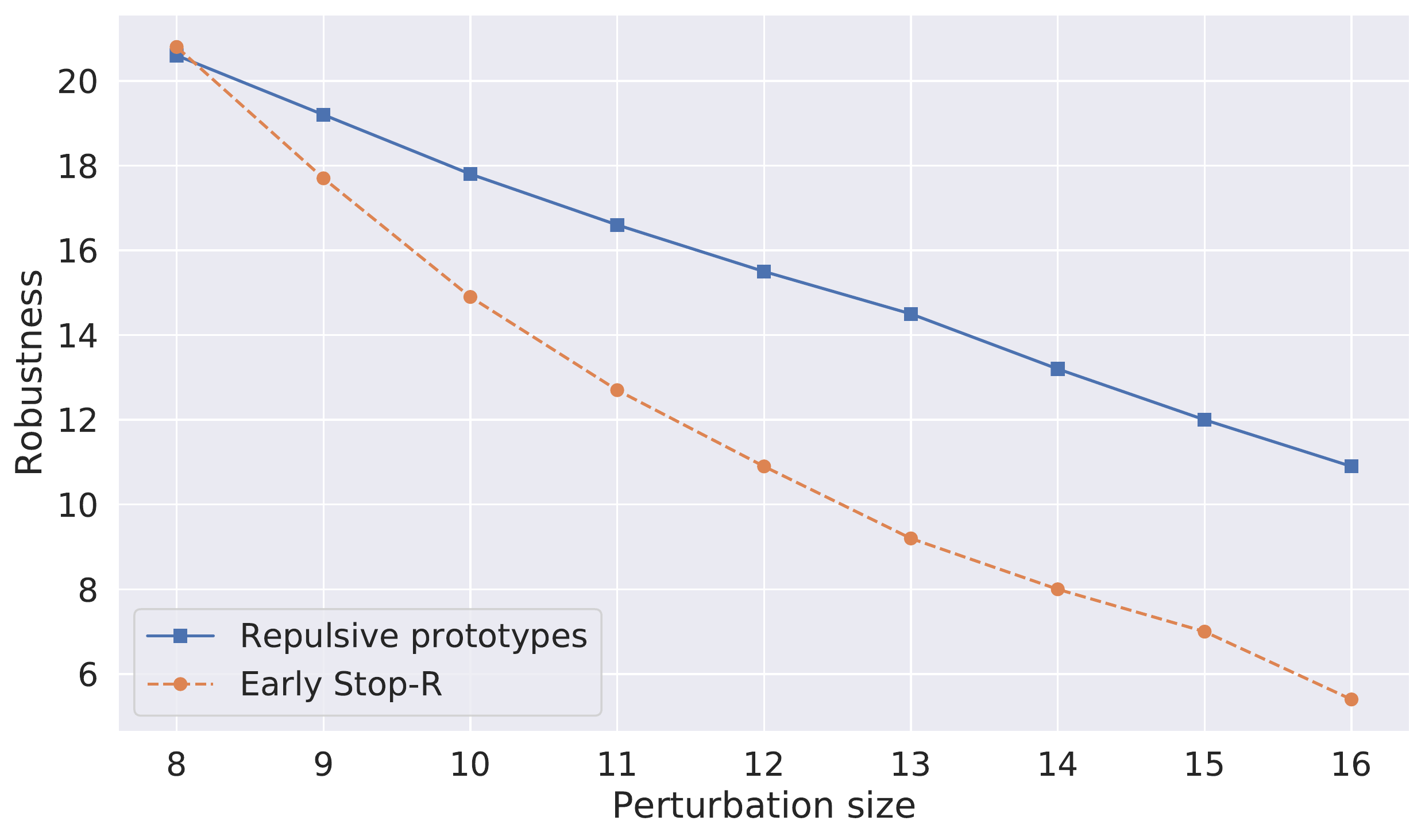}
    \caption{Robustness curves for CIFAR-100, obtained by testing with PGD-20, and various perturbation sizes ($\epsilon$).}
    \label{fig:cifar100_curve}
\end{figure}

Similarly to CIFAR-10, we present in Figure~\ref{fig:cifar100_curve} the robustness curve obtained by testing with different perturbation sizes.
We compare the results with the Early Stop-R model, which uses the final parameters published by~\citeauthor{rice2020overfitting}.
% Although the results could not be reproduced with precision, we are now interested in how fast robustness degrades when increasing the perturbation size.
We observe that, as for CIFAR-10, training with repulsive prototypes yields models resilient to large perturbation sizes, preserving more than half of the initial robustness when increasing the perturbation by a factor of two.
% However, models trained using adversarial training are less resilient to large perturbation sizes.

\subsection{Black-box evaluation}
\label{subsec:transfer}

Besides the white-box threat model investigated above, we evaluate the models in a black-box scenario.
In particular, we use the transferability attack, in which an attacker trains a substitute model and uses it to craft adversarial examples.

Black-box attacks are used to evaluate the model's robustness, but also to detect if the defences employed give a false sense of security~--~\eg~due to obfuscating gradients~\cite{athalye2018obfuscated}.
Since training with repulsive prototypes does not add any transformation or randomisation which may have adverse effects (such as gradient obfuscation), we expect the defence to behave similarly to adversarial training~--~a defence known to have no side effects.
Therefore, we compare transferability on repulsive prototypes with transferability on adversarially trained models.

\citet{su2018robustness}~showed that the architecture can impact transferability.
Particularly when the network's building blocks are different (\eg~between the Inception architecture which uses different filter sizes and ResNet which uses invariant filter sizes and residual connections), robustness has higher variance.
However, when the building blocks are the same, but the depth of the network increases (\eg~ResNet-50 vs. ResNet-101), robustness has smaller variance.

Therefore, since all the models from this paper use a variant of the ResNet architecture, we use a substitute model based on it.
We also assume that an attacker has access to the training data set and does not need to apply data augmentation~\cite{papernot2016transferability}.
For crafting adversarial examples we use the Regular model from Tables~\ref{tab:cifar10} and \ref{tab:cifar100}. 
For testing, we use the \ac{pgd} attack with $n=20$
and compare with the Early Stop and Early Stop-R models. % from above.

\begin{figure}[t]
    \centering
    \includegraphics[width=7.5cm, keepaspectratio]{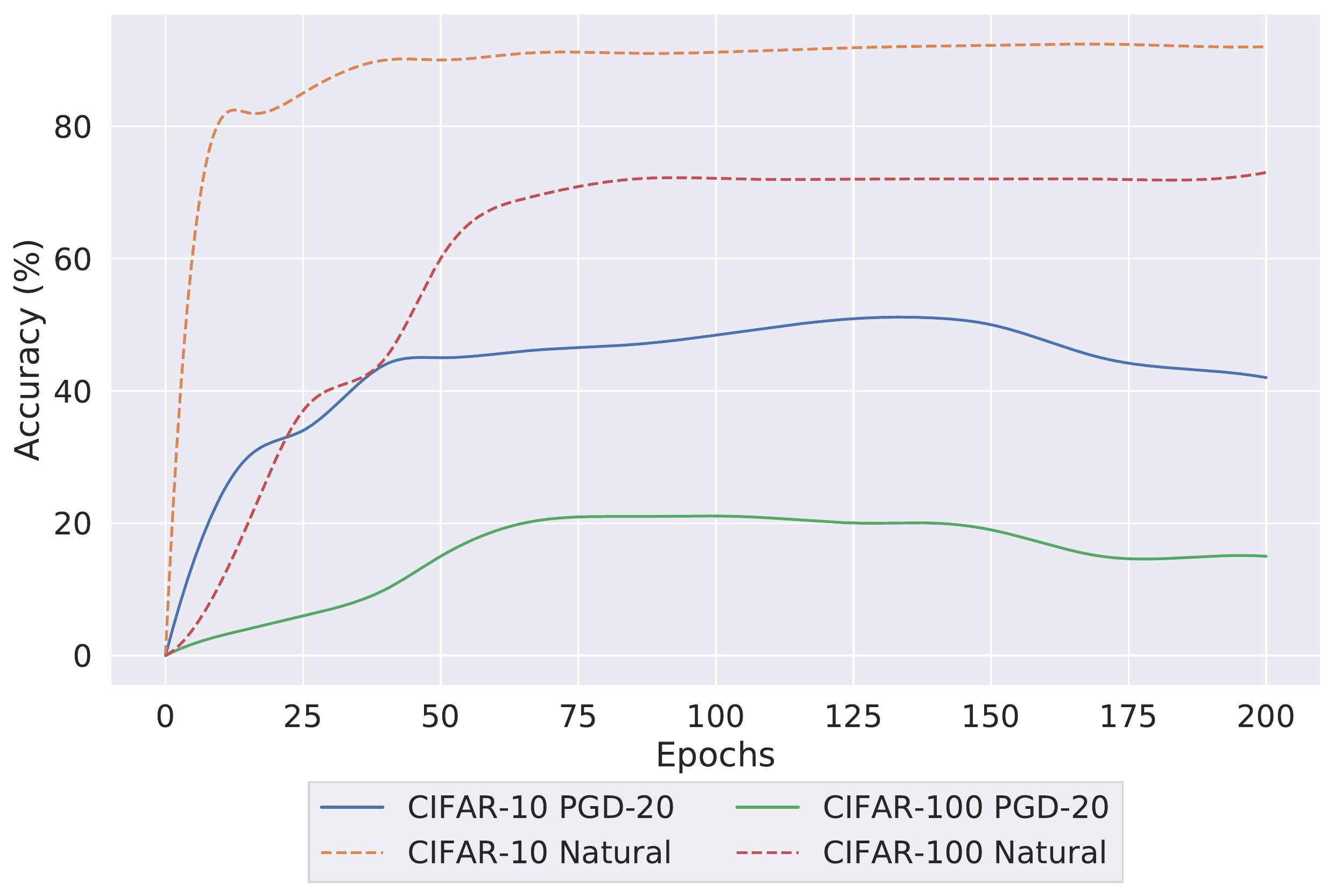}
    \caption{Overfitting in training with repulsive prototypes.}
    \label{fig:overfitting}
\end{figure}

\begin{table}[t]
    \centering
    \begin{tabular}{l|l|c|c}
         Source &  Target & CIFAR-10 & CIFAR-100 \\
        \toprule
        Regular\textsuperscript{1} & Repulsive\textsuperscript{1} &  72.5 & 36.2 \\
        Regular\textsuperscript{1} & Early Stop\textsuperscript{2} &  75.1  &  -  \\
        Regular\textsuperscript{1} & Early Stop-R\textsuperscript{3} &  -  &  37.4  \\        
        \bottomrule
    \end{tabular}
    \caption{Robustness against transferred adversarial examples. The models are \textsuperscript{1}ResNet-18, \textsuperscript{2}WideResNet-34-10, \textsuperscript{3}PreActResNet-18}
    \label{tab:transferability}
\end{table}

The results for both CIFAR-10 and CIFAR-100 are presented in Table~\ref{tab:transferability}.
We observe that (i) robustness against black-box attacks is higher than robustness against white-box attacks, which indicates that both defences have no adverse side effects such as obfuscated gradients, and (ii) both models achieve similar robustness, consistent with the results from Tables~\ref{tab:cifar10} and \ref{tab:cifar100}.
Since for CIFAR-10 Early Stop uses a more complex architecture, we expect the model to also have higher robustness (which corresponds with the results from~\citet{su2018robustness}, and the higher gap in Table~\ref{tab:transferability}).

 \begin{figure*}[t]
 	\centering
    \hspace*{-5.5ex}     	
	\subfloat[Repulsive.\label{fig:proto_cm}]{
	\adjustbox{valign=t}{\includegraphics[width=4.2cm, keepaspectratio]{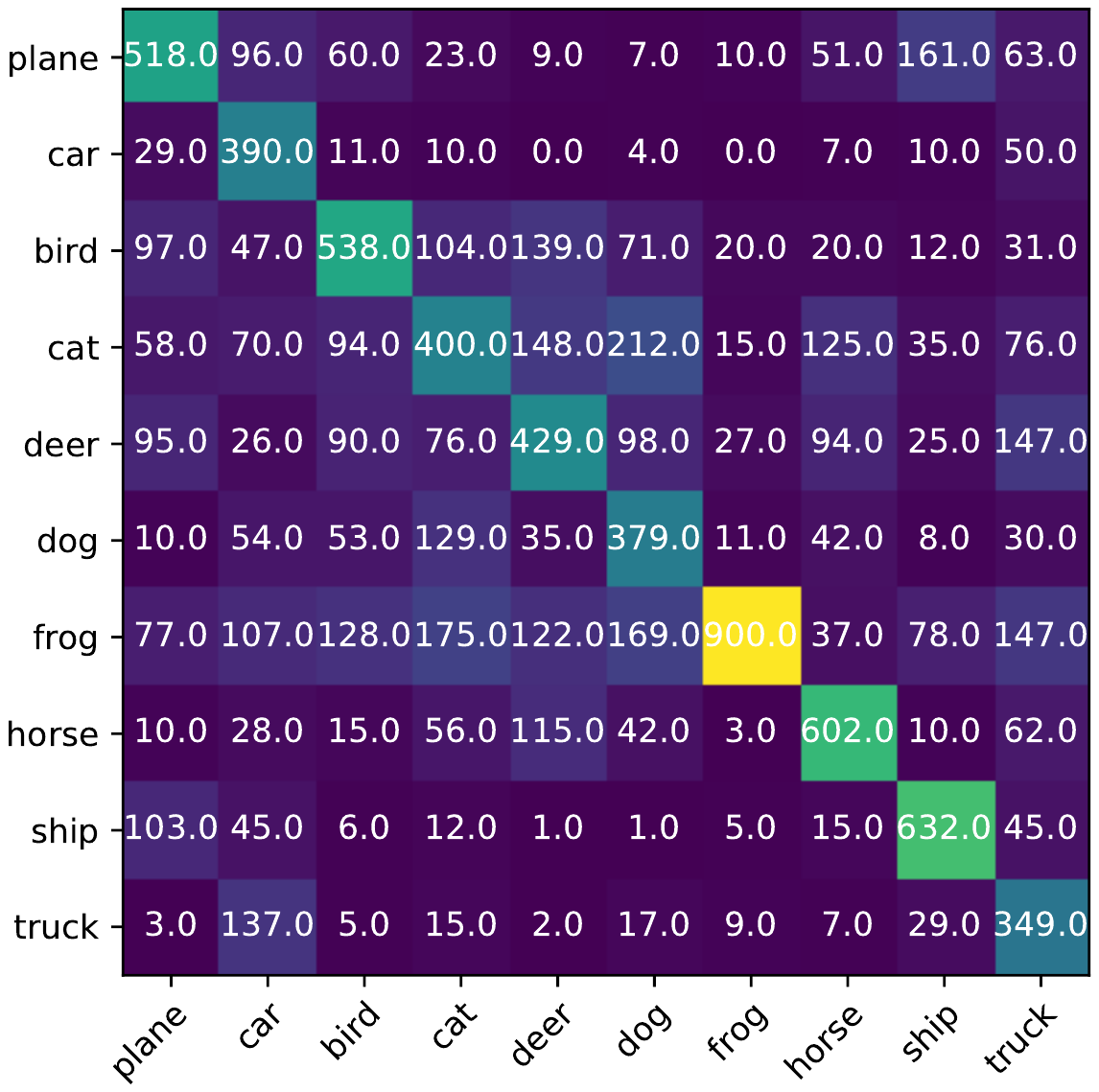}}}    
\hspace*{7.5ex}
   	\subfloat[Early Stop.\label{fig:ofit_cm}]{
   		\adjustbox{valign=t}{\includegraphics[width=4.2cm, keepaspectratio]{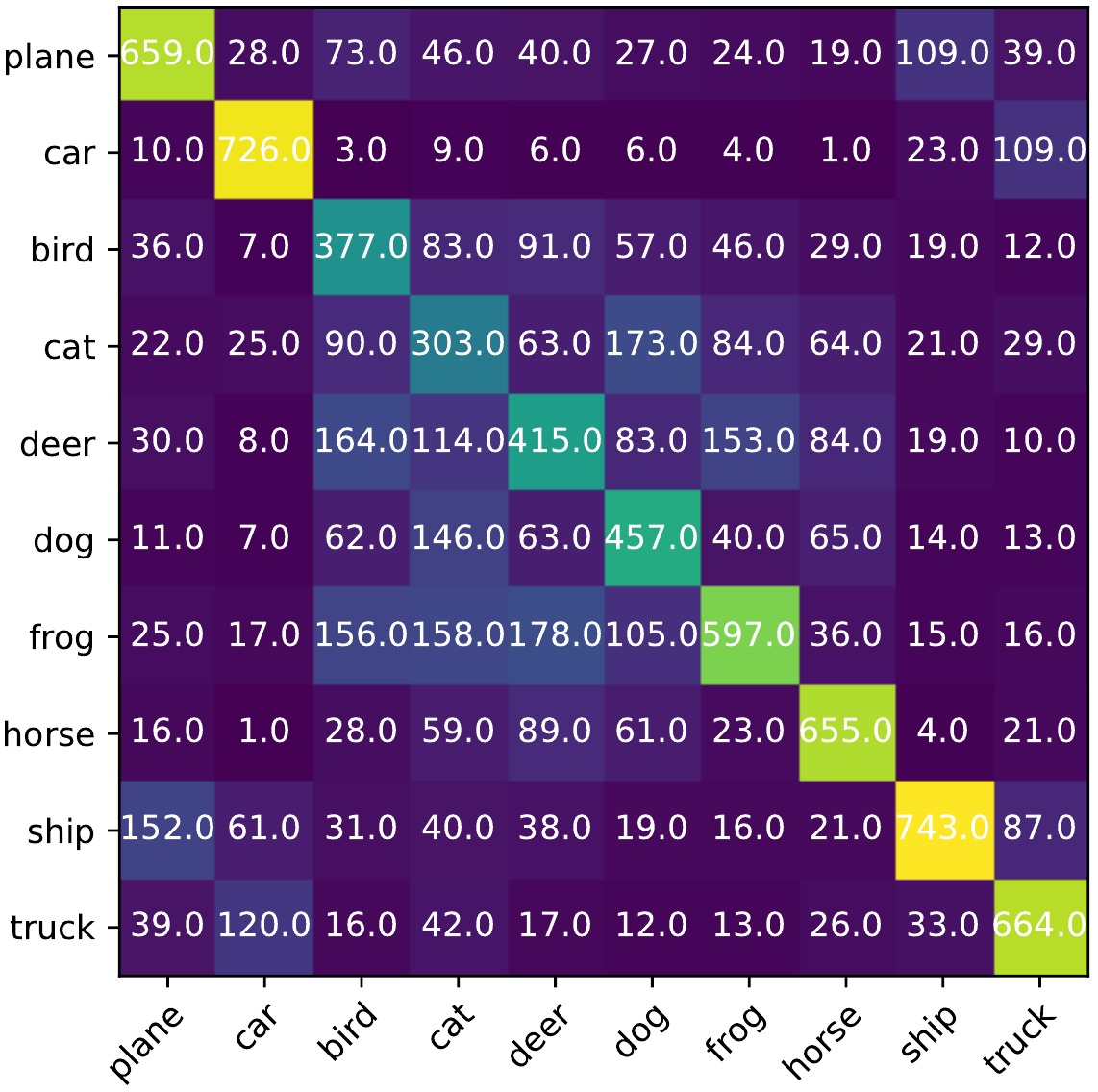}}}
 \hspace*{6.5ex}
 	\subfloat[Misclassified and nearest correctly classified samples.\label{fig:adv_ex_samples}]{
 		\adjustbox{valign=t}{\includegraphics[width=4.25cm, keepaspectratio]{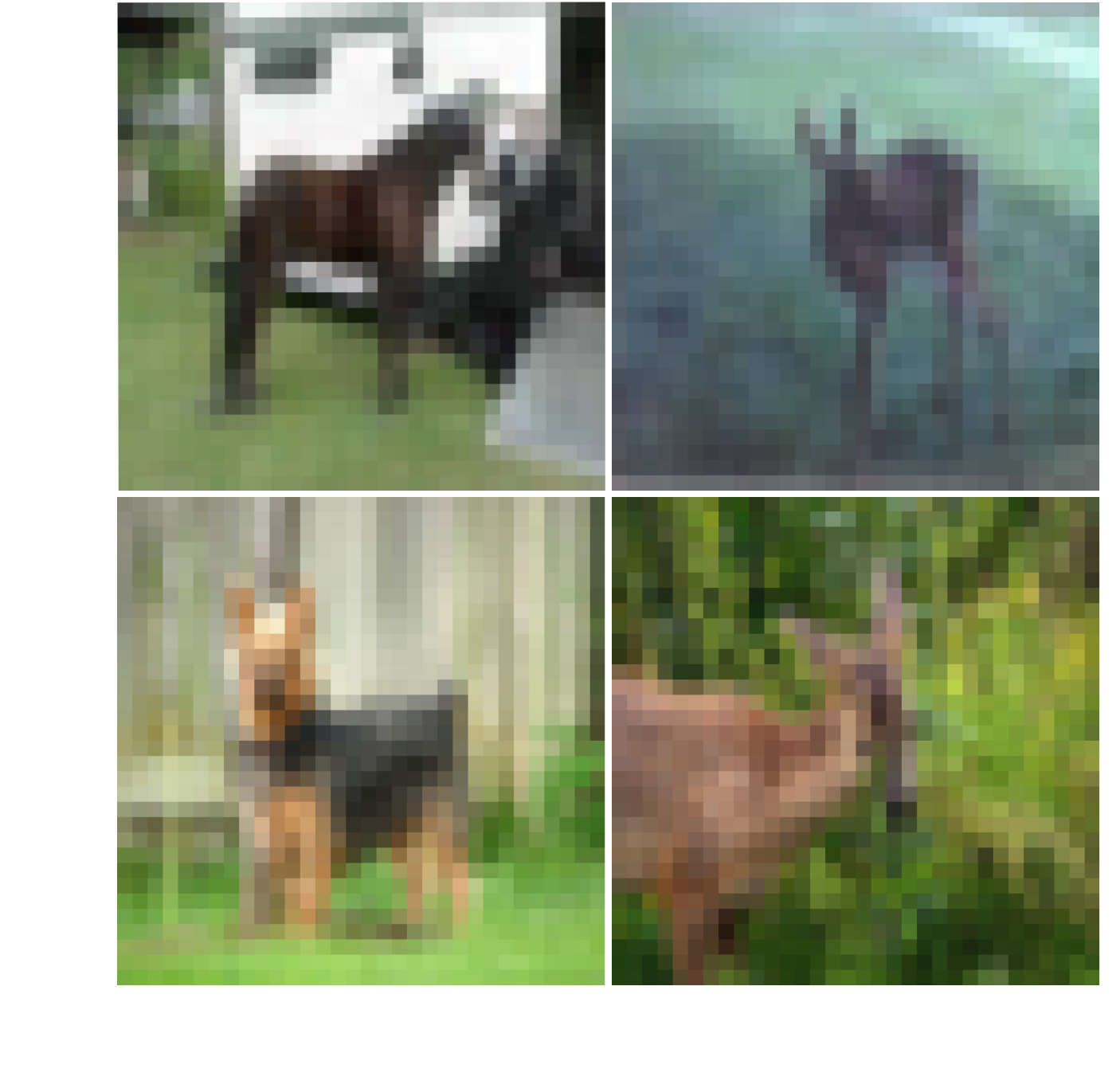}}}
 \caption{Adversarial confusion matrices for the (a) Early Stop, and for the (b) Repulsive models in Table~\ref{tab:cifar10}. Figure (c) shows misclassified samples (left) and the nearest correctly classified samples in the representation space (right).}
 \label{fig:cms}
 \end{figure*}

However, since the target models have different architectures and loss functions, we expect them to also have distinct internal representations.
Therefore, the transferability results should have higher variance than those from Tables~\ref{tab:cifar10} and~\ref{tab:cifar100}.
Yet the results from Table~\ref{tab:transferability} suggest that the target models have similar failing modes,
and raise the question if some samples are more sensitive to perturbations than others, and if the samples are common between the target models.
An enquiry follows in the next section.

\section{Discussion}
\label{sec:discussion}

Firstly, we investigate if robustness is linked to properties of the testing data set, or of certain samples.
Following the observation from the last section~--~that models trained with repulsive prototypes and with adversarial examples behave similarly against black-box attacks~--~we plot the confusion matrices for adversarial examples on CIFAR-10, for the Repulsive and Early Stop models (Figures~\ref{fig:ofit_cm} and~\ref{fig:proto_cm}).

We observe that 60\% of the top-1 misclassified classes are the same for both models (\eg~planes misclassified as ships).
This percentage increases to 90\% when we judge the top-2 classes (\eg~planes misclassified as ships or birds).
We also analysed the overlap between misclassified adversarial examples by the two models and found that over 65\% of the samples were common.
% We also analyses the overlap between the misclassified adversarial examples by the Early Stop model and by the Repulsive model and found that that 69\% of the misclassified adversarial samples are the same.
This result indicates that some samples may be more sensitive to perturbations.
For the Repulsive model, it indicates that for some samples training fails to provide intra-class compactness. 
When perturbed, these samples are easier to move to incorrect regions.

In order to further investigate this phenomenon, we performed random sampling on the misclassified examples for manual inspection.
For all samples, we also extracted the closest examples from the predicted (wrong) class.
Two such examples are displayed in Figure~\ref{fig:adv_ex_samples}, where the pictures on the left are the incorrectly classified examples and the ones on the right are the closest examples in the predicted class.
We observe that both examples have many common characteristics with the closest sample in the incorrect classification regions.
Similar results could be observed for other samples (suppressed due to space constraints).
% Particularly for the first example, where given the angle of the shots, the horse and the deer have similar physical appearances.
Previously, \citet{jo2017measuring}~showed that neural networks have a tendency to learn surface regularities rather than higher-level abstractions.
Our initial investigation indicates that samples with similar surface regularities are also more sensitive to adversarial perturbations, even if the models using them are trained with distinct loss functions.
A deeper investigation into this phenomenon is planned for future work.

Secondly, we  note that training with repulsive prototypes for adversarial robustness is more sensitive than pursuing the highest accuracy on natural samples.
This is a consequence of the loss function (\eqq~\ref{eq:loss}) which measures a distance, and taking large steps towards minimising it may push samples closer to the classification boundaries, rather than closer to the prototypes.
Larger steps are not relevant for natural samples, but are important for robustness.
In order to alleviate this effect we used smaller learning rates and early stopping.
However, this also means that models trained with repulsive prototypes are prone to overfitting for robustness.
Figure~\ref{fig:overfitting} shows the behaviour of the Repulsive models trained on both data sets studied.
We observe that, while the models are stable on natural samples, they are prone to overfitting against adversarial examples.
% Moreover, the models are more prone to overfitting on repulsive prototypes than in adversarial training~\cite{rice2020overfitting}.
Moreover, the models seem likely to overfit faster when trained with repulsive prototypes than with adversarial training~\cite{rice2020overfitting}.

Lastly, we note that prototypes designed prior to training can also embed other properties in the output space.
For example, \citeauthor{mettes2019hyperspherical} used word2vec \cite{mikolov2013distributed} to designed prototypes.
% For example, the structure from word2vec~\cite{mikolov2013distributed} can be used to design prototypes with privileged information~\cite{mettes2019hyperspherical}.
Adding more structure to the output space may lead to higher abstractions~--~\eg~to compositionality, as in word2vec~--~and it is an interesting avenue for future work.

\section{Conclusions and Future Work}
\label{sec:conclusion}
% TODO: Describe ablation studies?
We introduce deep repulsive prototypes for adversarial robustness~--~a training method which partitions the output space prior to training into prototypes with large class separation, and train models to preserve it.
Repulsive prototypes help models to gain robustness competitive to adversarial training, while removing the need to generate adversarial examples.
Moreover, models trained with repulsive prototypes are less sensitive to large perturbations and trade less accuracy on natural samples for robustness.

Our results indicate that the output space size is important for robustness, and that test samples with similar surface regularities are more sensitive to adversarial perturbations.
For future work we propose to search for better ways to design prototypes for robustness, which may embed other properties to the output space than large inter-class separation.
Moreover, we plan to further investigate if misclassified samples present similar surface regularities with samples in the predicted class, and find ways to remove the tendency of neural networks to rely on surface regularities.

\bibliography{bibliography}
% \bibliographystyle{aaai21}

% \appendix
% \section{Prototype selection CIFAR-100}
% \label{sec:protocifar100}

% \input{tables/proto_cifar_100}

\end{document}